\documentclass{article}

\PassOptionsToPackage{numbers}{natbib}

\usepackage[final]{neurips_wrl2021}

\usepackage[pdftex]{graphicx}
\usepackage{amsmath}
\usepackage{bm}
\usepackage[linesnumbered,ruled,vlined]{algorithm2e}
\SetKwInput{KwInput}{Input}                 
\SetKwInput{KwOutput}{Output}               

\DeclareMathOperator*{\argmin}{arg\,min}
\newcommand{\E}[2]{\mathbb{E}_{#1}\left[ #2\right]}

\newcommand{\x}{\bm{x}}

\newcommand{\z}{\bm{z}}

\usepackage{xcolor}

\usepackage[utf8]{inputenc} 
\usepackage[T1]{fontenc}    
\usepackage{hyperref}       
\usepackage{url}            
\usepackage{booktabs}       
\usepackage{amsfonts}       
\usepackage{nicefrac}       
\usepackage{microtype}      

\usepackage{wrapfig}
\usepackage{subcaption}

\makeatletter
\def\@mb@citenamelist{cite,citep,citet,citealp,citealt,citepalias,citetalias}
\makeatother
\usepackage[resetlabels,labeled]{multibib}
\newcites{A}{Appendix References}

\usepackage{xparse}
\let\oriCiteO\citepA

\RenewDocumentCommand{\citepA}{O{} O{} m}{%
  \renewcommand{\citenumfont}[1]{A##1}%
  \oriCiteO[#1][#2]{#3}%
  \renewcommand{\citenumfont}[1]{##1}%
}

\title{Versatile Inverse Reinforcement Learning via Cumulative Rewards}

\author{%
  Niklas Freymuth \thanks{correspondence to \texttt{niklas.freymuth@kit.edu}} \\
 \And
   Philipp Becker \\
   Autonomous Learning Robots\\
   Karlsruhe Institute of Technology\\
   Karlsruhe, Germany\\ \\
 \And
   Gerhard Neumann \\
}

\begin{document}

\maketitle

\begin{abstract}
Inverse Reinforcement Learning infers a reward function from expert demonstrations, aiming to encode the behavior and intentions of the expert.
Current approaches usually do this with generative and uni-modal models, meaning that they encode a single behavior.
In the common setting, where there are various solutions to a problem and the experts show versatile behavior this severely limits the generalization capabilities of these methods.
We propose a novel method for Inverse Reinforcement Learning that overcomes these problems by formulating the recovered reward as a sum of iteratively trained discriminators.
We show on simulated tasks that our approach is able to recover general, high-quality reward functions and produces policies of the same quality as behavioral cloning approaches designed for versatile behavior.
\end{abstract}

\section{Introduction and Related Work}
\label{sec:Introduction}
\begin{figure}[t]
\centering
    \begin{minipage}{0.23\textwidth}
    \centering
    \includegraphics[width=.88\textwidth]{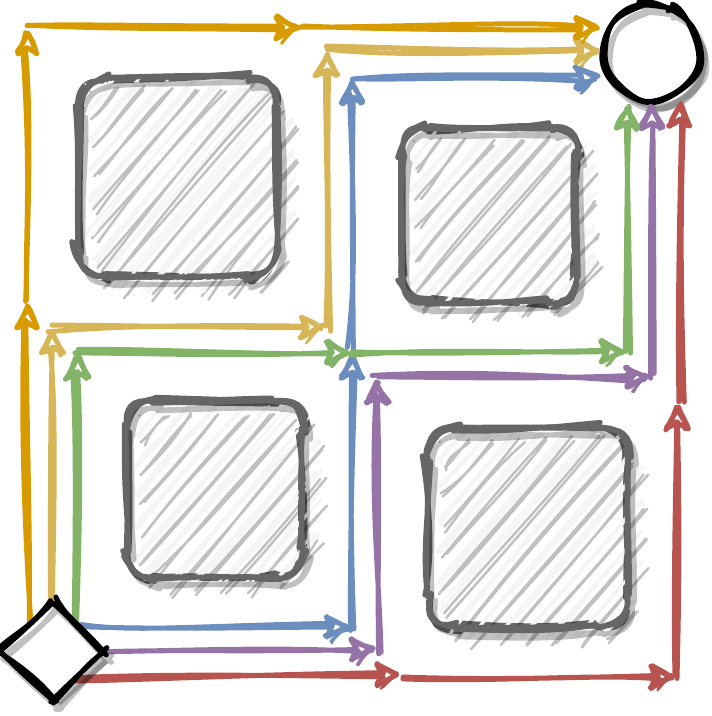}
    \subcaption[]{}
    \end{minipage}%
    \begin{minipage}{0.23\textwidth}
    \centering
    \includegraphics[width=.88\textwidth]{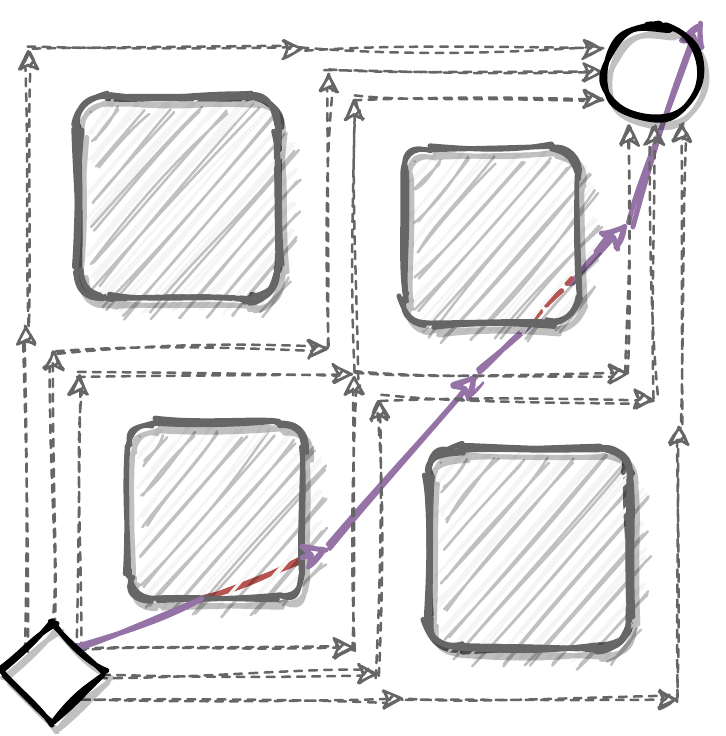}
    \subcaption[]{}
    \end{minipage}%
    \begin{minipage}{0.23\textwidth}
    \centering
    \includegraphics[width=.88\textwidth]{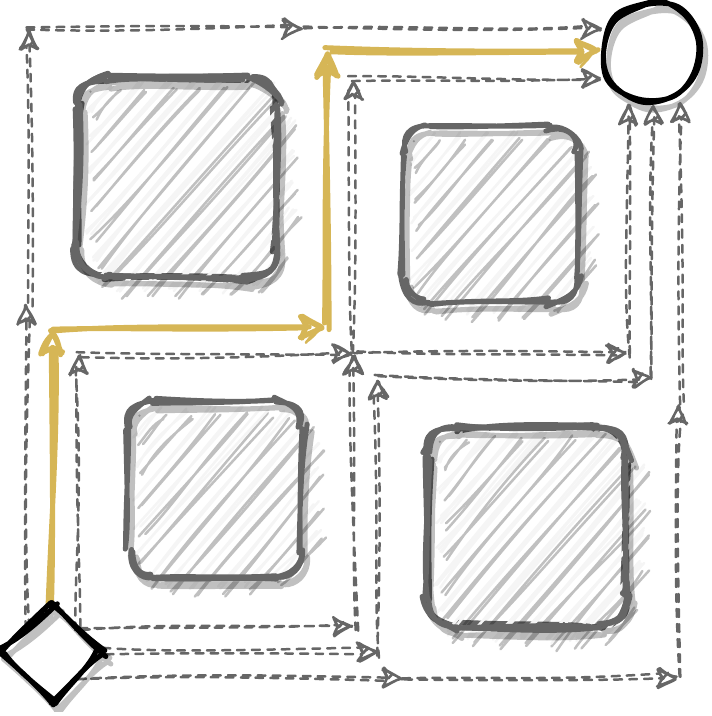}
    \subcaption[]{}
    \end{minipage}%
    \begin{minipage}{0.23\textwidth}
    \centering
    \includegraphics[width=.88\textwidth]{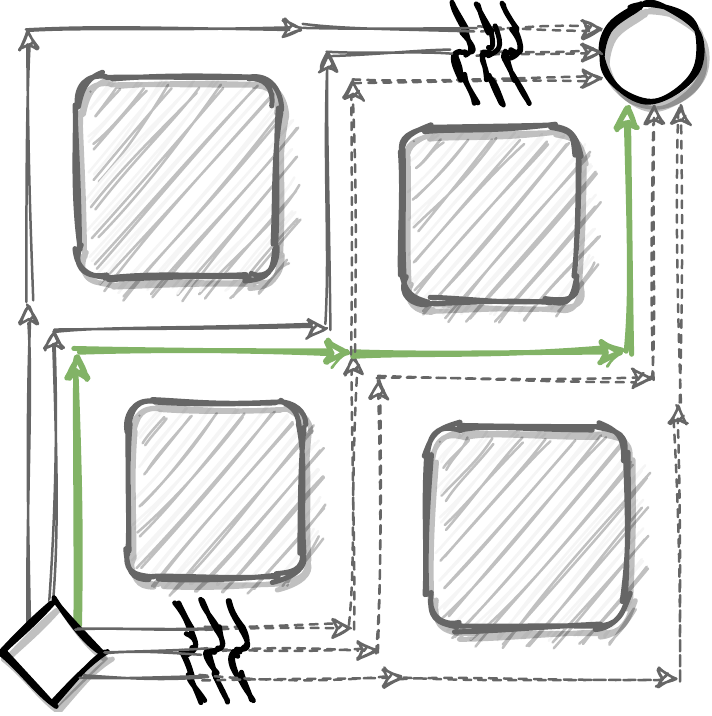}
    \subcaption[]{}
    \end{minipage}%
    \caption{\textbf{a)} A versatile path-panning task, where to goal is to find an efficient path from the diamond to the circle. \textbf{b)} Maximum likelihood methods average over modes, leading to poor solutions. \textbf{c)} Choosing a single behavior via the information projection works, but is not flexible. \textbf{d)} V-IRL reconstructs a versatile reward function that allows for good solutions even when some paths are obstructed.}
    \label{fig:introduction}
\end{figure}

Reinforcement Learning shows great potential for tasks with clearly specified objectives, such as games \citep{mnih2015human, silver2016mastering, openai2019dota}.
Yet, in many real-world scenarios, manually specifying a suitable reward is complicated and can lead to unintended behavior. 
Thus, Inverse Reinforcement Learning (IRL) \citep{russell1998learning} aims at recovering a reward from expert demonstrations instead.  
Many tasks allow various solutions and different experts may chose different approaches, leading to multi-modal, versatile demonstrations.
Properly capturing this versatility does not only better reflects the nature of the task but also naturally provides robustness to changes in the environment. 
See Figure \ref{fig:introduction} for an example.
Many approaches  \citep{ziebart2008maximum, ziebart2010maxcausalent, ratliff2006maximum, brown2019extrapolating, brown2020better} use uni-modal policies together with maximum likelihood objectives, 
which forces the model to average over modes of data that cannot be represented.
Another recent line of work \citep{finn2016guided, fu2018learning} is closely connected to Generative Adversarial Networks \citep{goodfellow2014generative}. 
They build on Generative Aderversarial Imitation Learning \citep{ho2016generative}, but reparameterize the discriminator by inserting the learned policy. It can be shown that this modified discriminator then estimates the density of the expert demonstrations, which can be used as a reward. 
These methods implicitly optimize the reverse KL-divergence allowing them to focus on a single behavior. 
See e.g., \citep{ghasemipour2020imitationLearning, arenz2020nonadversarial} for a detailed discussion.
While this prevents averaging over multiple potential solutions,
the uni-modal nature of the policy still prevents properly capturing the versatility. 

We introduce Versatile Inverse Reinforcement Learning (V-IRL), an approach 
designed specifically for learning reward functions form highly-versatile demonstrations.
We build on recent works in Varitional Inference \citep{arenz2018efficient, arenz2020trust} and most notably on Expected Information Maximization (EIM) \citep{Becker2020Expected}, a method for density estimation that trains a mixture model where each component is able to focus on a different mode of the given data.
EIM is able to reliably model versatile behavior, but does not provide a reward function.
We extend EIM to still produce a similar multi-modal policy while also making its reward function explicit, producing a fine-grained and versatile reward in the process.
We experiment on diagnostic tasks designed to have versatile solutions.
The results show that V-IRL is the only method that is able to consistently capture the multimodal nature of the tasks while recovering a fine-grained reward.
\par 

\section{Algorithm}
\label{sec:algorithm}
We consider sample trajectories $\x$ which follow an unknown distribution $p(\x)$ and are given by the expert. 
Under the common maximum entropy assumption \citep{ziebart2008maximum, haarnoja2017reinforcement} the expert's reward is given by
$R\left(\x\right) = \log p\left(\x\right) - c$
for some constant offset $c$. Recovering the log density of the unknown distribution $p\left(\x\right)$ thus also recovers a reward which explains the experts behavior.

Expected Information Maximization (EIM) \citep{Becker2020Expected} iteratively minimizes the reverse KL-Divergence $\text{KL}\left(q\left(\x\right)||~p\left(\x\right)\right)$ between a latent variable policy $q\left(\x\right) = \int q\left(\x|\z\right)q(\z)d\z$ and an unknown distribution $p\left(\x\right)$ of which only samples are available. 
Similar to EM \citep{dempster1977maximum}, EIM uses a variational bound to make the optimization tractable. 
This bound is given by a reformulation of the bound used in \citep{arenz2018efficient}, $q_{t+1}^*\left( \x\right) =$
\begin{equation}
\label{eq:eim_objective}
 \argmin_{q\left(\x\right)}
\mathbb{E}_{q\left(\x|\z\right)q\left(\z\right)}
\left[ - \log \dfrac{p\left(\x\right)}{q^*_t\left(\x\right)}\right] + \text{KL}(q(\z) || q^*_t(\z)) + \mathbb{E}_{q\left(\z\right)}\left[\text{KL}\left(q\left(\x|\z\right) || q^*_t\left(\x|\z\right)\right)\right]\text{,}
\end{equation}
where $q^*_t\left(\x|\z\right)$ and $q^*_t\left(\z\right)$ denote the model from the previous iteration. 
Relating Equation \ref{eq:eim_objective} to IRL, $q_t\left(\x\right)$ can be seen as an iteratively optimized behavioral cloning policy. The KL penalties enforce that consecutive $q_t\left(\x\right)$ do not change too quickly, acting like trust regions \citep{schulman2015trust}.
To use this bound under the assumption that only samples of $p(x)$ are available, EIM employs density ratio estimation techniques \citep{sugiyama2012density} to approximate  $\log \left(p\left(\x\right) / q^*_t\left(\x\right)\right)$ with a neural network $\phi_t\left(\x\right)$. 
In the practical implementation of the approach the change of the model between iterations is limited and thus, in each iteration, the density ratio estimator can be reused after a few update steps. 

At iteration $t$, $q^*_{t+1}\left(\x\right)$ in Equation \ref{eq:eim_objective} is optimal for
$
    q^*_{t+1}\left(\x\right) \propto \exp \left(\log q^*_{t}\left(\x\right) +\phi_t\left(\x\right)\right)
$ \citep{abdolmaleki2015model}.
From there, induction over $t$ shows that
$
    \log q^*_{t+1}\left(\x\right) = \log q^*_{0}\left(\x\right) + \sum_{i=0}^t \phi_i\left(\x\right) + c
$
for some arbitrary prior $q^*_{0}\left(\x\right)$ and constant offset $c$.
Assuming convergence at iteration $T$, plugging this into $R\left(\x\right) = \log p\left(\x\right) - c$ yields
\begin{equation}
\label{eq:cumulative_reward}
\begin{alignedat}{-1}
    R\left(\x\right) 
    = &\log q^*_0\left(\x\right) +\sum_{i=0}^{T-1} \phi_i\left(\x\right)\text{,}
\end{alignedat}
\end{equation}
where we dropped constant offsets as they do not play a role in optimization.
Intuitively, each $\phi_t\left(\x\right)$ acts as a change of reward that refines the current recovered reward $q^*_{t}\left(\x\right)$.
By adding more $\phi_t\left(\x\right)$s, this gradually produces a more and more precise estimate of the reward. At convergence, the target distribution $p\left(\x\right)=q^*_T\left(\x\right)$ is recovered and $\phi_T\left(\x\right)=0$ everywhere.
Since each $q^*_t\left(\x\right)$ is the accumulation of $t$ different estimators $\phi_i\left(\x\right)$, we call this a \textit{cumulative reward}.

Comparing V-IRL to EIM, we see that V-IRL can represent an arbitrarily complex reward, as each additional $\phi_t\left(\x\right)$ adds capacity to the model. Opposed to this, EIM only recovers a policy and is thus limited to the capacity of this policy by construction.
V-IRL also acts in an off-policy setting, allowing for large update steps of the policy without destabilizing the training. In contrast, both EIM and generative approaches \citep{finn2016guided, fu2018learning} often need to take sufficiently small steps for the method to converge \citep{arenz2020nonadversarial}.
Formulating the IRL problem as a sum of changes of rewards is conceptually easier than having a generative reward, because estimating a ratio is easier than estimating a density \citep{sugiyama2012density}. 
Another benefit lies in the multi-modal structure of our approach. Since the log density ratio estimate $\phi_t\left(\x\right)$ gives a strong signal in areas of uncovered expert modes by design, the recovered reward will eventually cover the relevant modes of the expert distribution and naturally prefer those with higher density over lower density ones.
If some modes become unavailable at inference time due to changes in dynamics caused by e.g., obstructions, a policy trained on the recovered reward can still use the remaining ones as a reward signal.

\paragraph{Importance Sampling}
In general, $q^*_t\left(\x\right)$ may become arbitrarily complex and can not easily be samples from.
To work around this, we introduce a tractable sampling policy $\tilde{q}_t\left(\x\right)$ that approximates $q^*_t\left(\x\right)$.
In our experiments, we use a recent method for variational inference, Variational Inference by Policy Search (VIPS) \citep{arenz2018efficient, arenz2020trust} to train $\tilde{q}_t$ by minimizing $\text{KL}\left(\tilde{q}_t\left(\x\right)||q^*_t\left(\x\right)\right)$.
Given $\tilde{q}_t\left(\x\right)$, we employ importance sampling to train a discriminator between expert demonstrations and samples of $\tilde{q}_t\left(\x\right)$ that are weighted by $q^*_t\left(\x\right)$. These samples then act as the importance sampling estimate of samples of $q^*_t\left(\x\right)$. To do this, we minimize the weighted binary cross entropy loss
\begin{equation}
    \label{eq:i2rl_phi_objective}
    \mathcal{L}_{BCE}(\phi_t, p, \tilde{q}_t, q^*_t) =
    \E{\x\sim p\left(\x\right)}{-\log \sigma\left(\phi_t\left(\x\right)\right)}+\E{\x\sim \tilde{q}_t\left(\x\right)}{-w_t\left(\x\right)\log \left(1-\sigma\left(\phi_t\left(\x\right)\right)\right)}
\end{equation}
with normalized importance weights 
$w_t\left(\x\right) = \frac{1}{\int q^*_t\left(\x\right)\mathop{\text{d}\x}} \frac{q^*_t\left(\x\right)}{\tilde{q}_t\left(\x\right)}$ and logits $\phi_t\left(\x\right)$. It has been shown that Equation \ref{eq:i2rl_phi_objective} causes the logits $\phi_t\left(\x\right)$ of the network to recover a log density ratio estimate at convergence \citep{sugiyama2012density}. In other words, $\phi_t\left(\x\right) = \log \left(p\left(\x\right) / q^*_t\left(\x\right)\right)$, which is precisely the change of reward used in Equation \ref{eq:cumulative_reward}.
Finally, we add the newly obtained $\phi_t\left(\x\right)$ to $q^*_t\left(\x\right)$ to obtain a new reward estimate. We iterate over this until convergence at iteration $T$, and obtain $q^*_T\left(\x\right)$ as the recovered reward and $\tilde{q}_T\left(\x\right)$ as an optimal policy for this reward.
Pseudocode for the approach can be found in Appendix \ref{app_sec:pseudocode}.

\paragraph{Kernel Density Estimation}

\begin{wrapfigure}{r}{0.53\textwidth}
\vspace{-16pt}
\begin{minipage}[b]{0.23\textwidth}
        \centering
    \includegraphics[width=\textwidth]{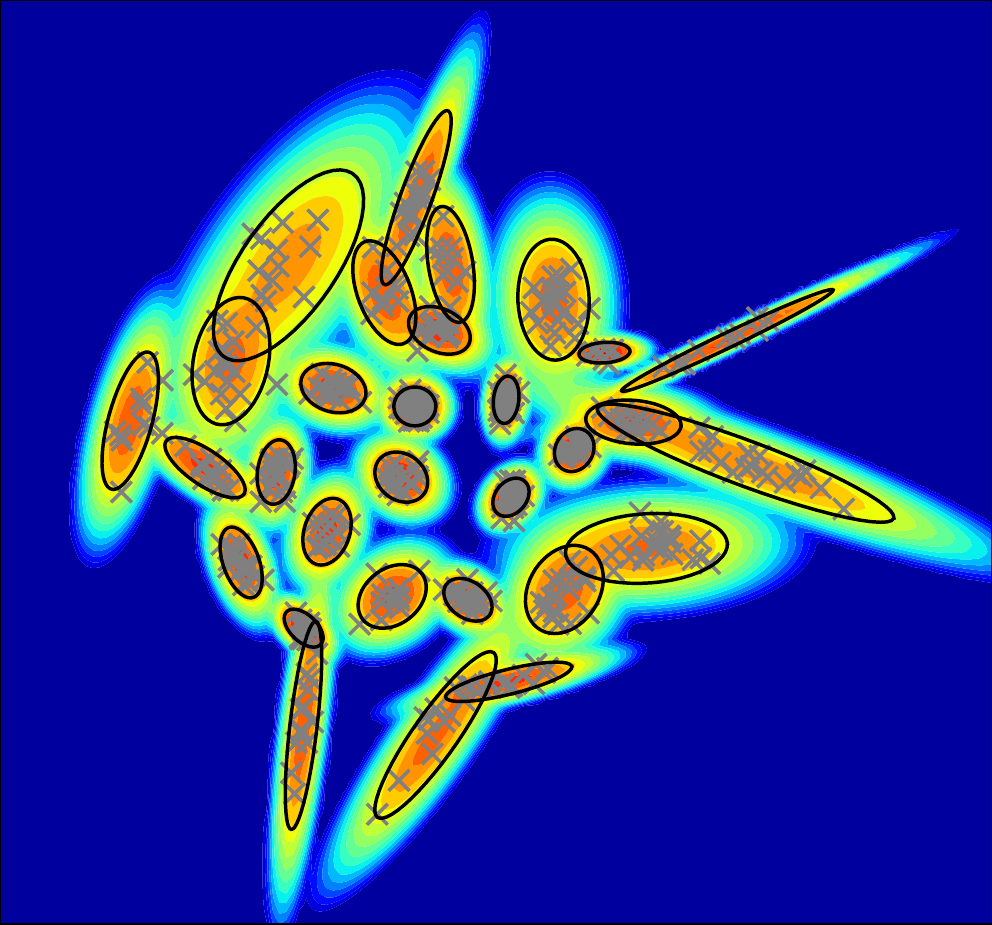}
    \subcaption{}
\end{minipage}%
\begin{minipage}[b]{0.3\textwidth}
    \centering
    \includegraphics[width=\textwidth]{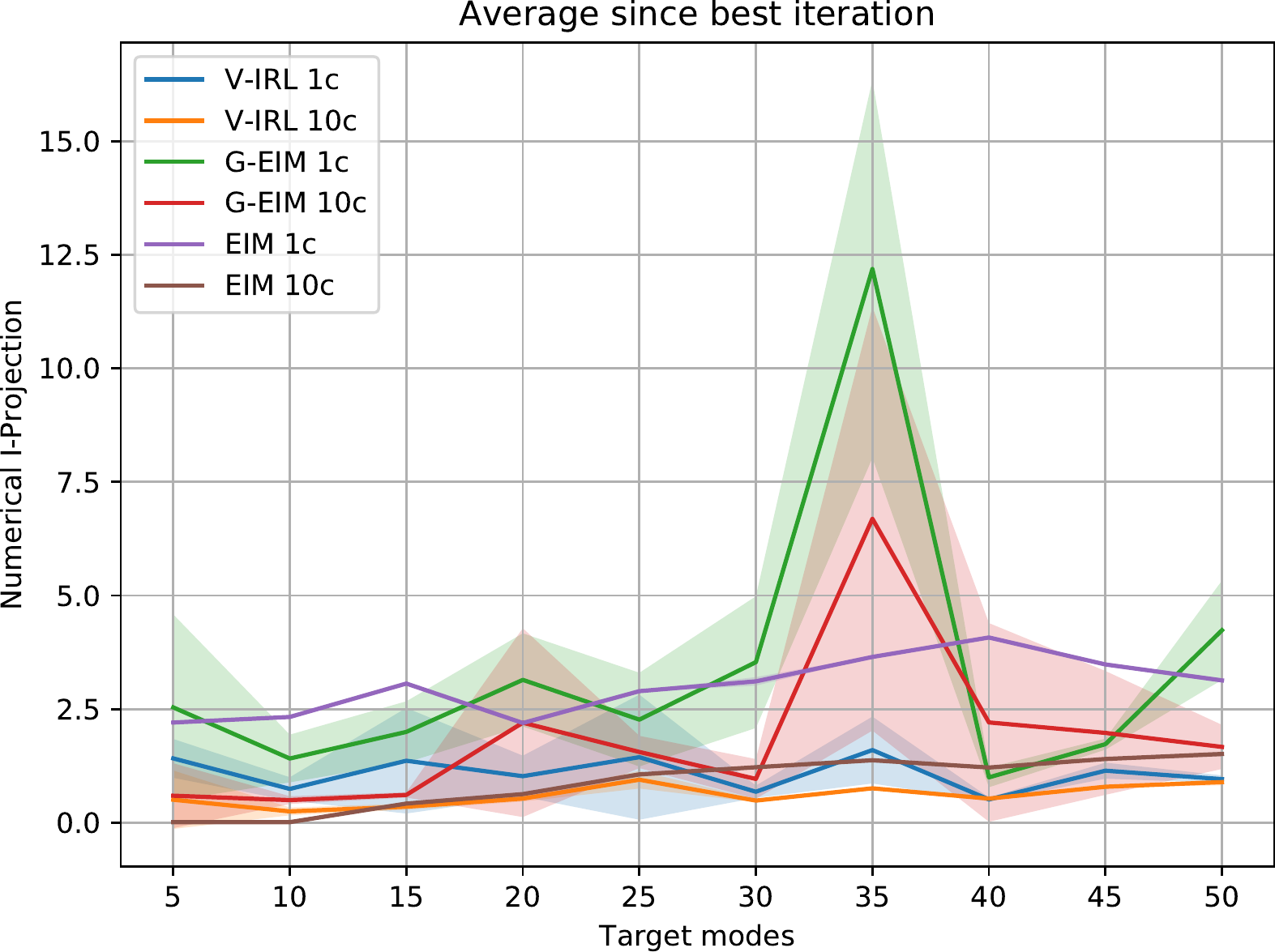}
    \subcaption{}
\end{minipage}
\caption{
    \textbf{a)}: Visualization of a $30$-component GMM. The heatmap shows the log-density of the GMM. The ellipsoids are $95\%$-covariances of the components. Expert demonstrations are shown with a grey `$\times$'.
    \textbf{b)}: Results for different numbers of target components.
    V-IRL provides a stable solution and is able to represent a larger number of modes than the baselines for the same number of policy components.
}
\label{fig:gaussians}
\end{wrapfigure}

\label{ssec:kernel_density_estimate}
In practical settings with highly-versatile behavior, $\tilde{q}\left(\x\right)$ is unlikely to cover all relevant modes of $q^*\left(\x\right)$, causing high-variance estimates in the importance sampling procedure and thus a poor coverage of the density ratio estimators and subsequently the reward recovered from them.
To prevent this, we follow prior work \citep{finn2016guided, fu2018learning} and enrich the sampling distribution $\tilde{q}_t\left(\x\right)$ with an estimate $\tilde{p}\left(\x\right)$ of the target distribution $p\left(\x\right)$, resulting in a fusion distribution 
$
    \mu_t\left(\x\right) = 0.5\tilde{q}_t\left(\x\right)+0.5\tilde{p}\left(\x\right)\text{.}
$
\citet{finn2016guided} and \citet{fu2018learning} use a Gaussian for $\tilde{p}\left(\x\right)$.
Since this is ill-suited for versatile behavior, we instead resort to a Kernel Density Estimate \citep{rosenblatt1956, parzen1962estimation} of the expert samples, using a Gaussian kernel for our estimate.
Thus, $\mu_t\left(\x\right)$ covers all modes of $p\left(\x\right)$ by construction. Intuitively, $\tilde{p}\left(\x\right)$ has the purpose of roughly covering $p\left(\x\right)$ to stabilize the training process, while $\tilde{q}_t\left(\x\right)$ is used to explore and approximate some modes of $p\left(\x\right)$ more closely.

\section{Experiments}
\label{sec:experiments}
We conduct all our experiments in a non-contextual and episodic setting. We choose Gaussian Mixture Models (GMMs) for the policies of the different methods. 
Note that extensions to a contextual setting are readily available in the form of Gaussian Mixtures of Experts, which we leave as a promising direction for future work. For hyperparameter settings, refer to Appendix \ref{app_sec:hyperparameters}.
We use EIM \citep{Becker2020Expected} as a baseline for versatile behavioral cloning, using the log-density of its policy as a surrogate reward function. Additionally, we adapt Adversarial Inverse Reinforcement Learning (AIRL) \citep{fu2018learning} to versatile tasks by combining it with EIM.
We use the EIM objective of Equation \ref{eq:eim_objective} for the general training procedure, but reparameterize the density ratio according to the AIRL formulation as
 $\phi_t\left(\x\right) = \log \left(\exp(R_t\left(\x\right)) / q_t\left(\x\right)\right)$.
Here, $q_t\left(\x\right)$ is the density of the current sampling policy and $R_t\left(\x\right)=\log p\left(\x\right)$ is a learned function that recovers the reward up to a constant. We call this approach generative EIM (G-EIM), and also enrich it with the fusion distribution mentioned in Section \ref{ssec:kernel_density_estimate}.
Note that G-EIM is a novel and interesting approach for versatile IRL in itself.

\paragraph{Gaussian Experiments}

We start with a versatile toy task to showcase the ability of VIRL to precisely capture highly multi-modal distributions. For this task, the reward is represented by the log-density of a $2$-dimensional GMM with $m$ randomly drawn and weighted components, leading to a complex and highly multi-modal target reward. An example for $m=30$ is given in the left of Figure \ref{fig:gaussians}. 
We evaluate the reverse KL by computing the numerical I-Projection near the ground truth data, comparing EIM, G-EIM and V-IRL with $1$ (`1c') and $10$ (`10c') GMM components each. 
We optimize for $m=50$ target components and evaluate on $m\in\{5, 10, \dots, 50\}$ components. We report the average results between the best evaluated iteration and the final iteration to account for instabilities in the training and evaluation. Results for $5$ random seeds can be seen in the right of Figure \ref{fig:gaussians}. We find that V-IRL scales gracefully with the number of modes to be covered, comparing favorably to both baselines. Inference experiments on the recovered rewards can be found in Appendix \ref{app_sec:additional_experiments}.

\paragraph{Grid Walker}

\begin{figure}[t]
\centering
    \begin{minipage}[b]{0.23\textwidth}
    \centering
    \includegraphics[width=\textwidth]{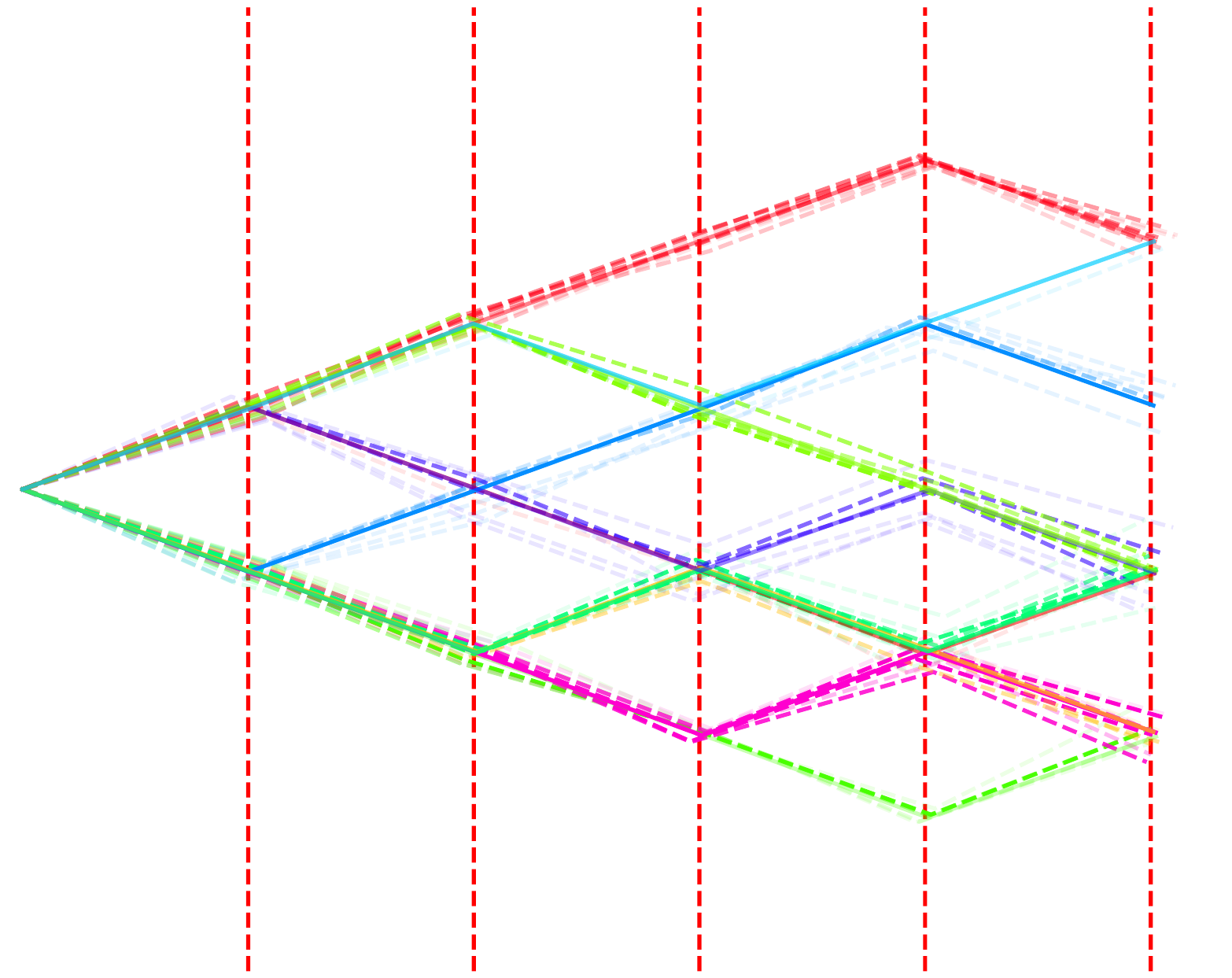}
    \subcaption[]{G-EIM Training}
    \end{minipage}%
    \begin{minipage}[b]{0.23\textwidth}
    \centering
    \includegraphics[width=\textwidth]{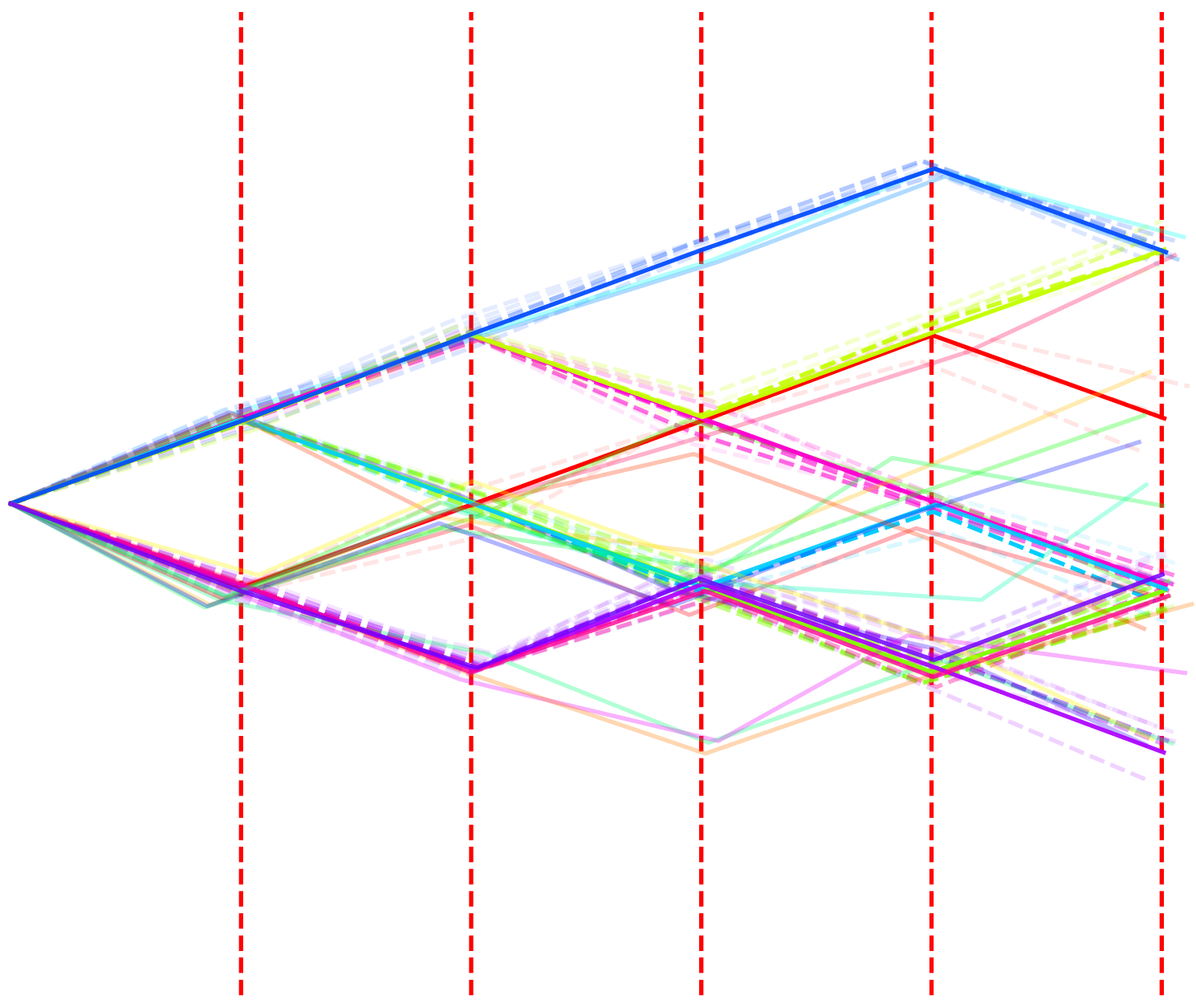}
    \subcaption[]{G-EIM Inference}
    \end{minipage}%
    \begin{minipage}[b]{0.23\textwidth}
    \centering
    \includegraphics[width=\textwidth]{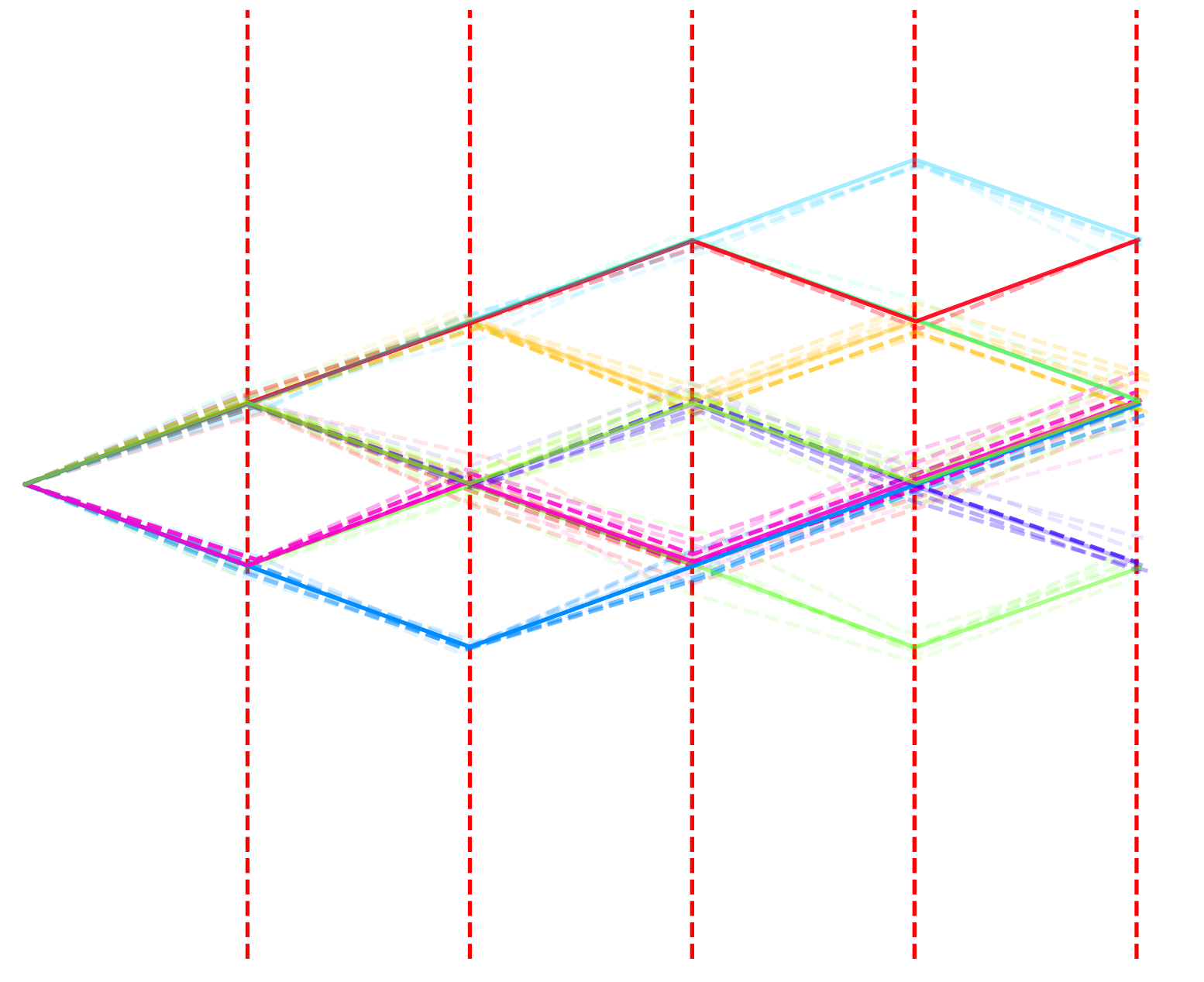}
    \subcaption[]{V-IRL Training}
    \end{minipage}%
    \begin{minipage}[b]{0.23\textwidth}
    \centering
    \includegraphics[width=\textwidth]{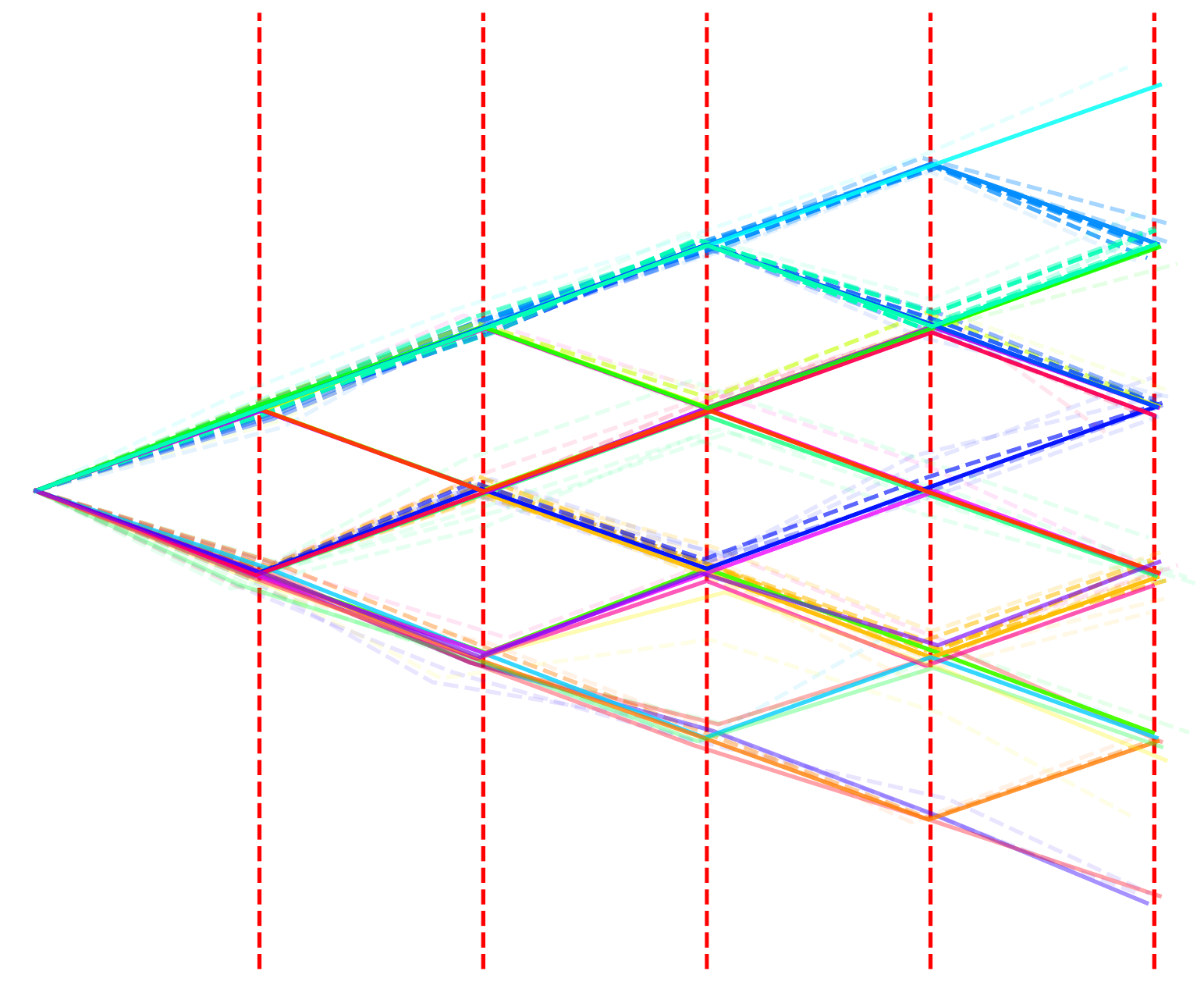}
    \subcaption[]{V-IRL Inference}
    \end{minipage}
    \caption{Grid walker experiments. Each walk is made up of $d=5$ line segments. The solid lines are policy component means, the dotted lines are samples from these components. Lines with a higher opacity correspond to a higher evaluation of the recovered reward. 
    The intermediate targets are denoted by the vertical dotted lines.
    \textbf{a)} a $10$-component sampling policy of G-EIM. \textbf{b)} a $25$-component inference policy trained on the reward recovered by a).
    \textbf{c)} and \textbf{d)} repeat this for V-IRL.
    Both V-IRL and GEIM produce high-quality sampling policies during training. 
    However, only V-IRL is able to also recover a reward function that extends to modes that were not seen during training.}
    \label{fig:inference_grid_walker_vis}
\end{figure}

A more realistic highly-versatile task is the path-planning task of Figure \ref{fig:introduction}. We model this task as a walk over $d$ steps of uniform size, where each step is given by an angle. For optimal solutions, the walker can either go up or down in each step.
As a result, all efficient solutions lie on a regular grid.
Appendix \ref{app_ssec:grid_walker} presents a detailed construction of the task.
We train V-IRL and G-EIM with $10$-component policies for $d=5$ path segments, using the negative ELBO (see e.g., \citep{blei2017variational}) as the optimization metric. The resulting hyperparameters can be seen in Table \ref{tab:app_optuna_hyperparameters_walker}. We then perform inference on the recovered rewards for policies with $25$ components. The results are shown in Figure \ref{fig:inference_grid_walker_vis}.
We find that the sampling policies for both V-IRL and G-EIM are versatile and precise. However, the inference policy of G-EIM is mostly limited to behaviors that were found during the training of the recovered reward.
Opposed to this, the recovered reward of V-IRL allows for previously unexplored modes to be found by the inference policy. We explore this behavior in more detail in Appendix \ref{app_ssec:grid_walker}.

\section{Conclusions}
We propose a novel approach for Inverse Reinforcement Learning for versatile behavior that recovers a reward by accumulating iteratively trained discriminative models. The key idea of our approach is that every discriminator is trained to represent an optimal change of the previous sum of discriminators, and thus this sum ultimately recovers an appropriate reward. We show that this cumulative reward formulation works well in a variety of versatile tasks, outperforms strong baselines and yields rewards that generalize beyond the capabilities of the sampling policy.

\subsubsection*{Acknowledgments}
The authors acknowledge support by the state of Baden-Württemberg through bwHPC.

\bibliographystyle{plainnat}
\bibliography{bibliography}

\newpage
\appendix
\section{Pseudocode for V-IRL}
\label{app_sec:pseudocode}

\setcounter{algocf}{-1}  
\begin{algorithm}[t]
\SetAlgoLined
\DontPrintSemicolon
\caption{V-IRL}
\KwInput{Expert Samples $\chi_{p}=\{\x_p^{\left(i\right)}\}_{i=1\dots M}$}
\KwInput{Reward prior $q^*_0\left(\x\right)$}
\KwInput{Initial sampling distribution $\tilde{q}_0\left(\x\right)$}
\KwOutput{Reward $R\left(\x\right) = \log p\left(\x\right)+c$}
\KwOutput{Optimal policy $\tilde{q}\left(\x\right)$ for this reward}
\For{$t=0\dots$}{
Train $\phi\left(\x\right)$ on $\left(\tilde{q}_t, q^*_t, \chi_{p}\right)$ using weighted binary cross-entropy \tcp*{Eq. \ref{eq:i2rl_phi_objective}}
Set $q^*_{t+1}\left(\x\right) = q^*_t\left(\x\right) \cdot \exp\left(\phi\left(\x\right)\right)$\\
Update $\tilde{q}_{t+1}\left(\x\right)$ by minimizing $\text{KL}\left(\tilde{q}_t\left(\x\right)||q^*_t\left(\x\right)\right)$ using e.g., VIPS
}
\KwRet{$\log q^*_{t+1}\left(\x\right)$, $\tilde{q}_{t+1}\left(\x\right)$}
\caption{Versatile Inverse Reinforcement Learning}
\label{alg:raw_i2rl}
\end{algorithm}

Algorithm \ref{alg:raw_i2rl} provides pseudocode for V-IRL.

\section{Additional Experiments}
\label{app_sec:additional_experiments}

\begin{figure}[h!]
    \centering
\includegraphics[width=.55\linewidth]{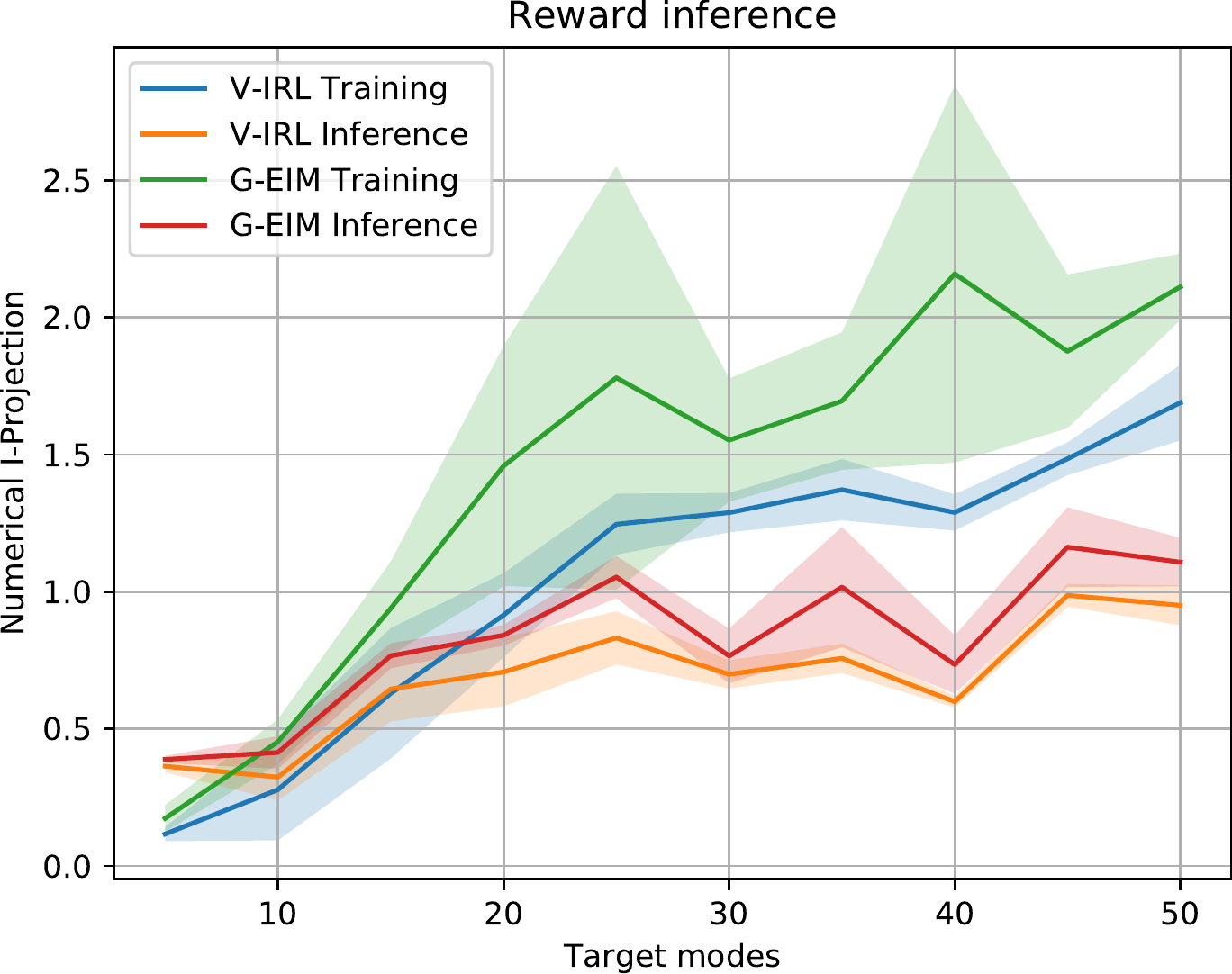}
    \caption{Numerical I-projection of inference policies trained on the best recovered rewards for V-IRL on the random Gaussian task. ‘Inference’ indicates VIPS with a number of components equal to the number of target modes trained on the final recovered reward. ‘Training’ refers to the 10-component learner policy.}
    \label{app_fig:gaussian_quantitative}
\end{figure}
\subsection{Random Gaussians}
\label{app_ssec:gaussian_experiments}
\begin{figure}
    \begin{minipage}{0.33\textwidth}
    \centering
        \includegraphics[width=.88\textwidth]{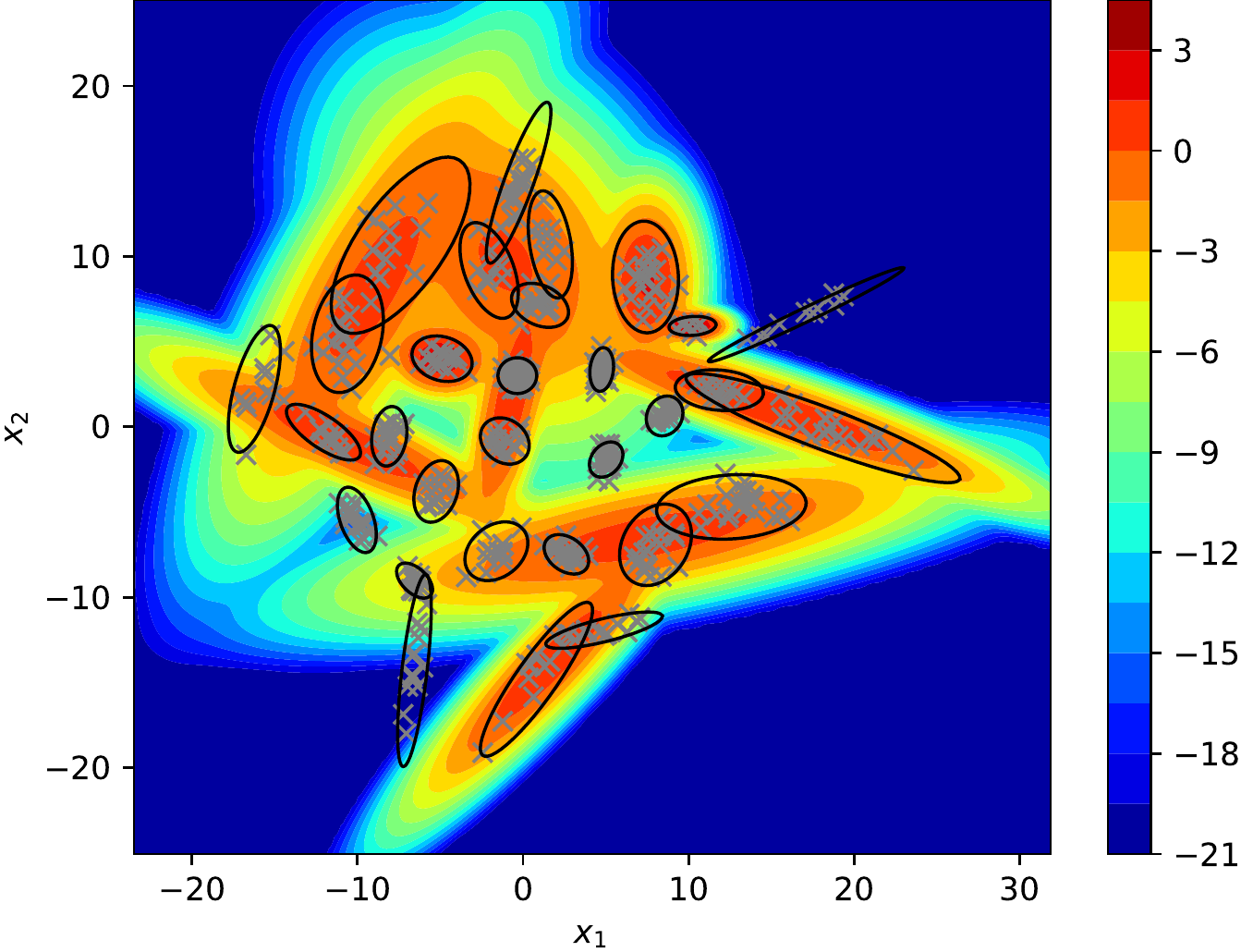}
        \subcaption[]{}
    \end{minipage}%
    \begin{minipage}{0.33\textwidth}
    \centering
        \includegraphics[width=.88\textwidth]{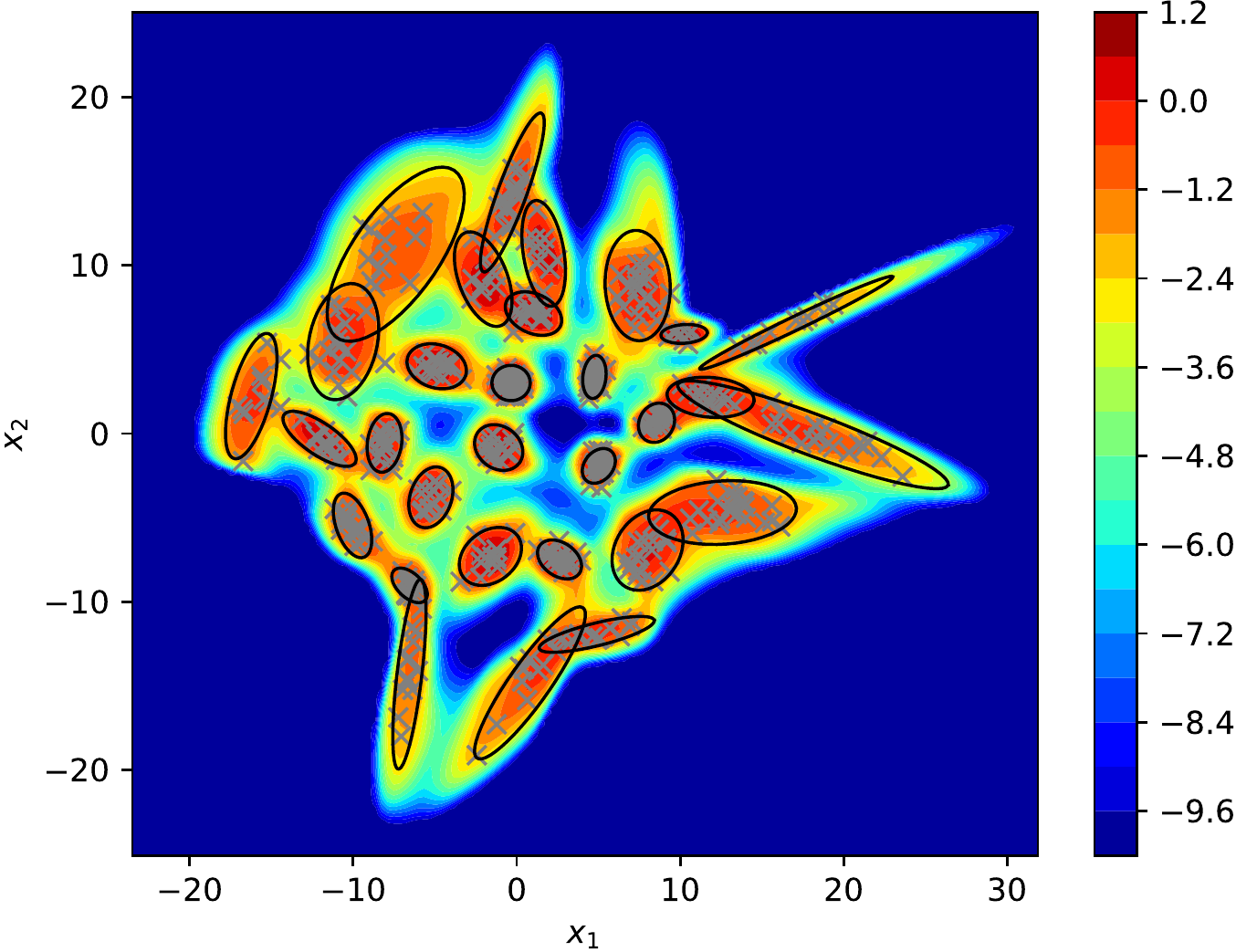}
        \subcaption[]{}
    \end{minipage}%
    \begin{minipage}{0.33\textwidth}
    \centering
        \includegraphics[width=.88\textwidth]{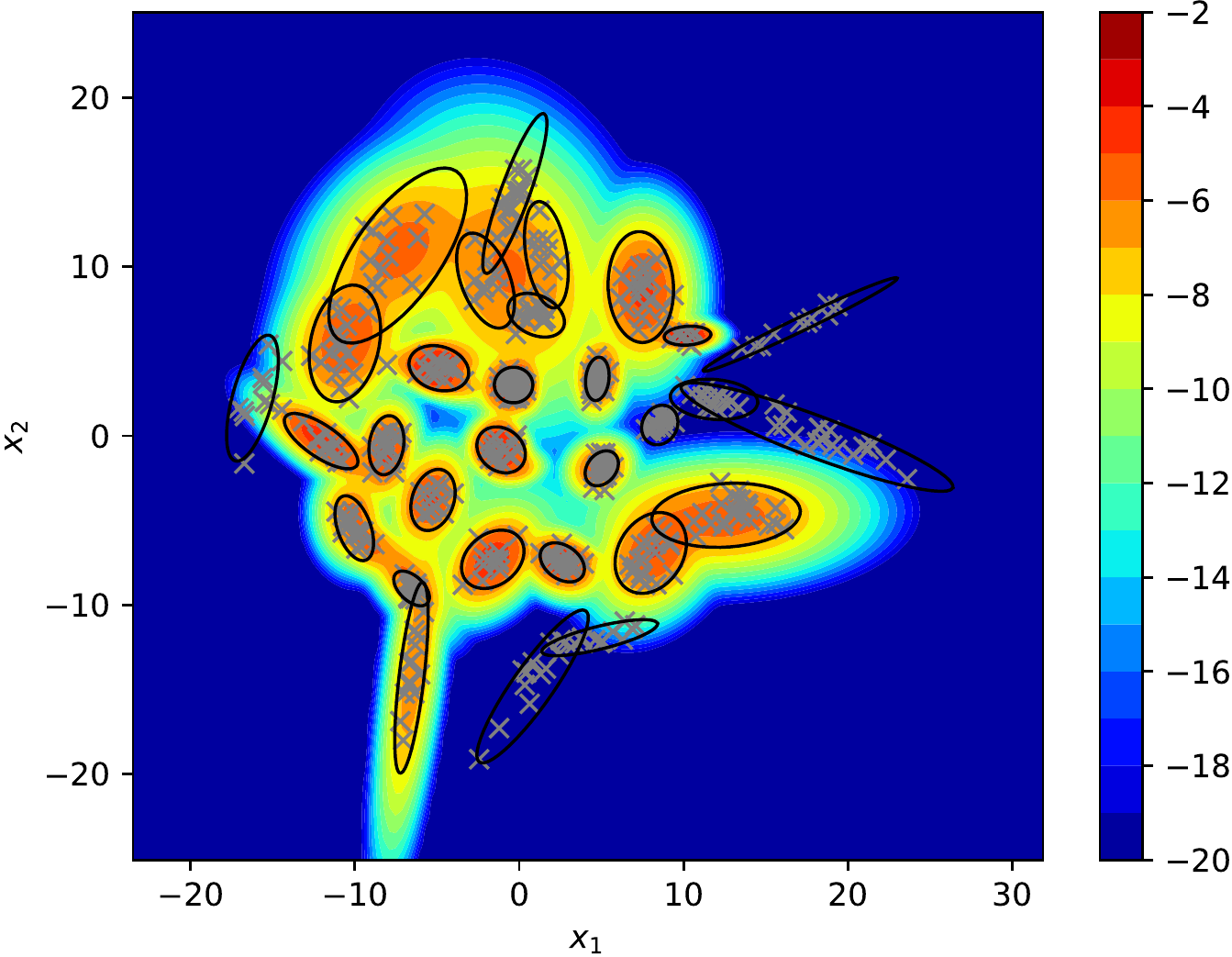}
        \subcaption[]{}
    \end{minipage}%

    \caption{
    Comparison of contour plots of \textbf{a)} the log-density of the sampling policy used by V-IRL, \textbf{b)} the average reward recovered by V-IRL over $5$ runs, and \textbf{c)} the log-density of a VIPS policy with $30$ components that is fit on the recovered reward of b).
    For all figures, the black ellipsoids are covariances of the individual expert components scaled to include $95\%$ of samples. $500$ randomly chosen expert samples are marked with a grey `$\times$'.
    }
    \label{app_fig:gaussian_qualitative}
\end{figure}

We use the reward recovered by the $10$-component versions of both V-IRL and G-EIM to train a newly initialized forward policy using VIPS with as many components as we have Gaussians in the task. 
We compare the numerical I-Projection between the $10$-component learner policies of the best reward iteration and the newly trained $m$-component policies evaluated at their best iteration in Figure \ref{app_fig:gaussian_quantitative}. Both methods generate a useful reward function that encodes more information than the policy that is used to produce it. However, V-IRL generally shows less variance in its solutions, and produces slightly better rewards overall.
A qualitative comparison between the sampling policy, the recovered reward and the inference policy for V-IRL is given in Figure \ref{app_fig:gaussian_qualitative}.

\par 

\subsection{Grid Walker}
\label{app_ssec:grid_walker}
\begin{figure}
    \centering
    \includegraphics[width=0.4\textwidth]{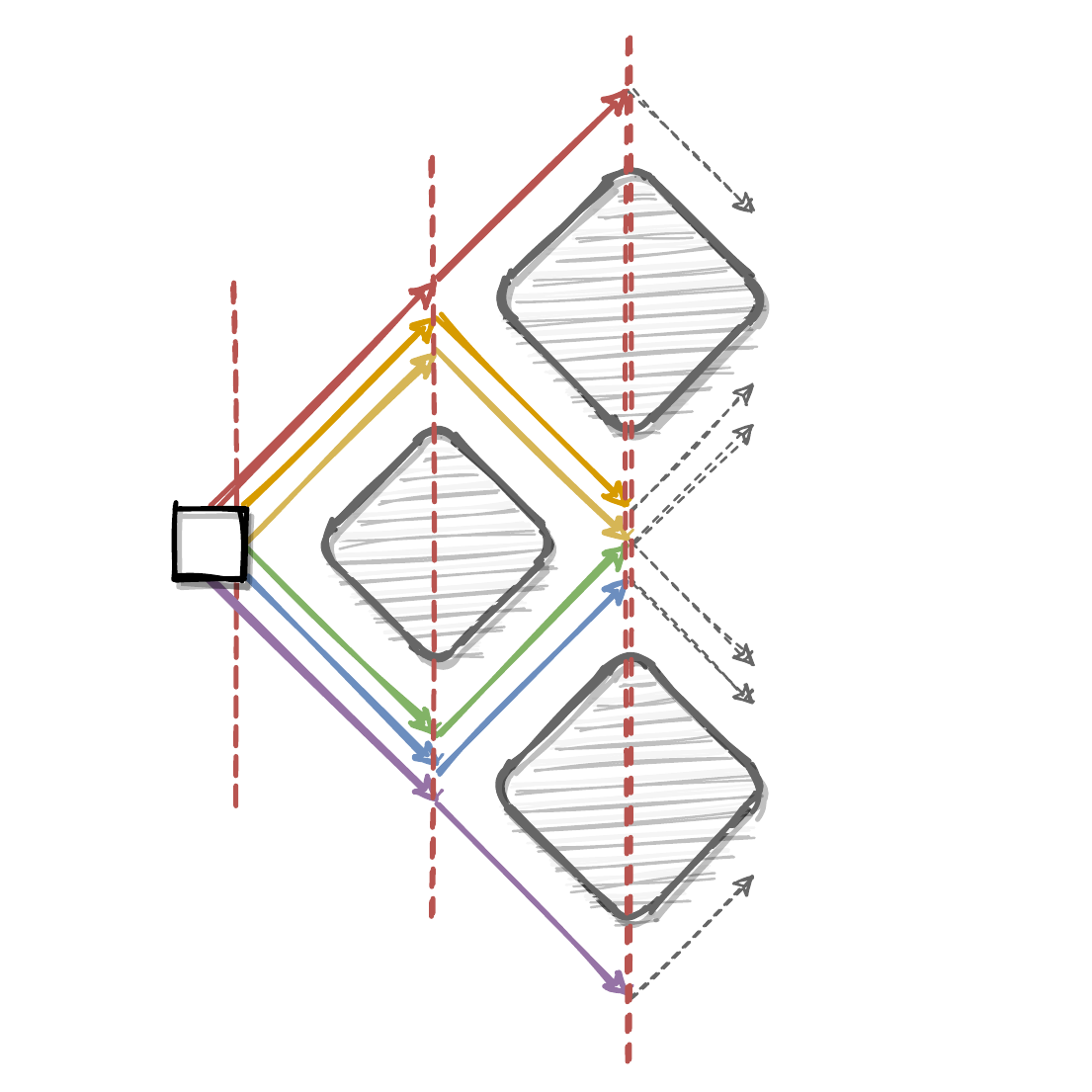}
    \qquad
    \includegraphics[width=0.5\textwidth]{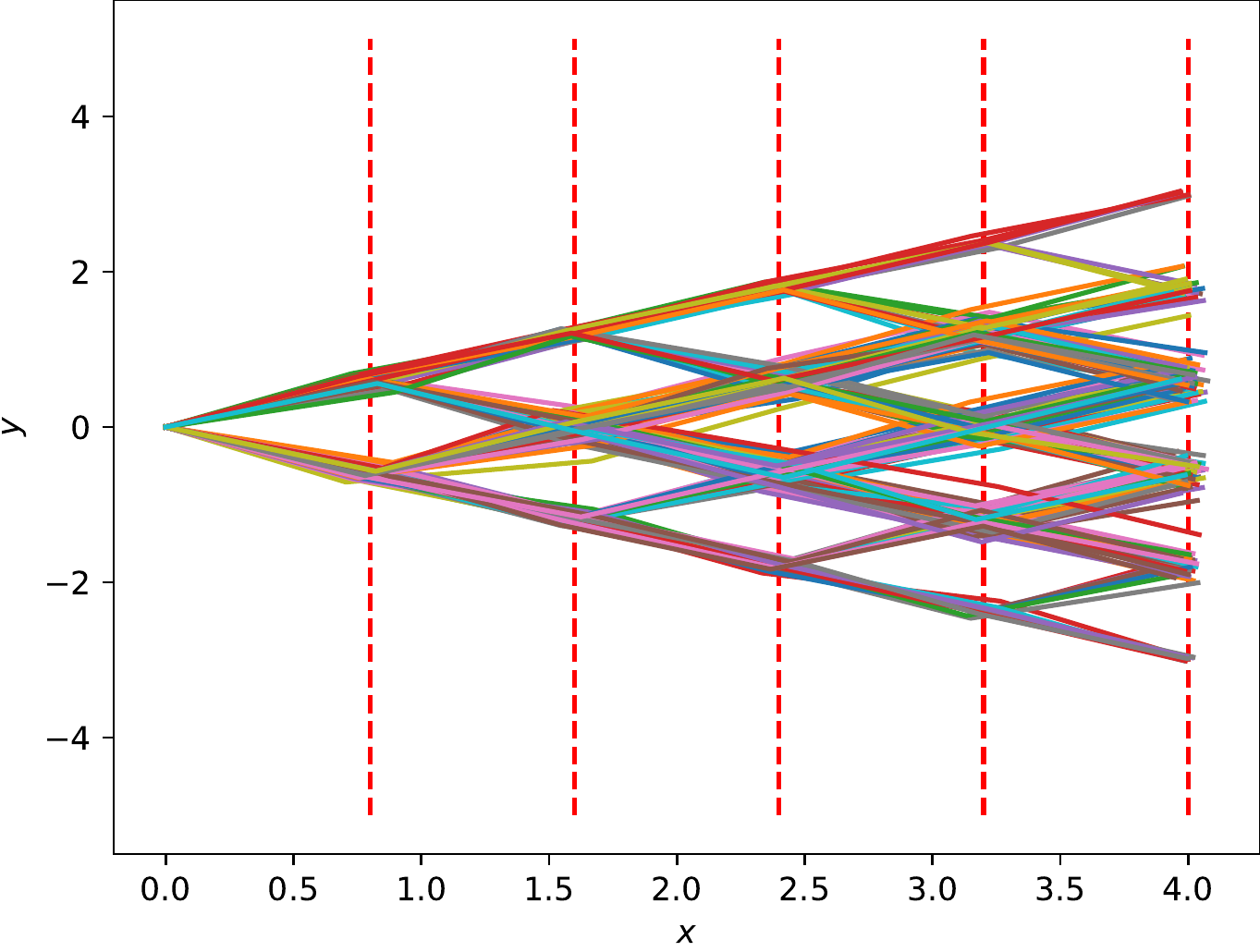}
    \caption{Grid walker construction. Left: Rotating the `first half' of the introductory example causes all efficient waypoints to lie on equidistant vertical lines. Identical steps share the same waypoints. Right: $200$ expert samples of the grid walker task for $d=5$ steps. All samples follow one of $2^5$ possible paths.
    These paths are implicitly encoded through $1$-dimensional Gaussians, which are visualized via the red dotted lines.}
    \label{app_fig:grid_walker_introduction}
\end{figure}

\paragraph{Construction}
We model the task as an agent that takes steps of length $1$ in a planar space, starting at $(0,0)^T$.
The action space is parameterized by an angle for each step, and the dimensionality of the task corresponds to the number of steps taken.
We only consider the first half of the path planning example of Figure \ref{fig:introduction} for our environment. Note that this encodes most of the versatility in the behavior of the agent.
To represent the choices of the agent more clearly, we rotate the example by $45^\circ$, aligning the waypoints along equidistant vertical lines. In this representation, all efficient paths lie on a grid inside a $90^\circ$ cone from the origin to the positive $x$-axis.
The result is visualized on the left of Figure \ref{app_fig:grid_walker_introduction}.

To construct the actual task, we define $\x$ to be the concatenation of absolute angles for the steps. That is, the position $h_i\left(\x\right)\in\mathbb{R}^2$ for a sample $\x=(x_1, \dots, x_d)^T$ at step $i$ is given by
\begin{equation*}
    h_i\left(\x\right) = h_{i-1}\left(\x\right) + \begin{pmatrix}
    \cos(x_i)\\ \sin(x_i)
    \end{pmatrix}
    =
    \begin{pmatrix} 
    \sum_{j=1}^i (\cos (x_j))\\ \sum_{j=1}^i (\sin (x_j)) \end{pmatrix}\text{,}
\end{equation*}
where $h_0\left(\x\right) = (0,0)^T$ for all $\x$.
We then introduce $d$ equidistant $1$-dimensional Gaussians centered on each line and calculate the likelihood of a walk as the product of the likelihood of each of its steps being drawn from its respective Gaussian.
This implicitly creates a grid-like structure, where for each segment the optimal behavior is to either `go up' or `go down'.
We set the distance of two consecutive Gaussians to $0.8$.
This causes the waypoints of a step to be positioned at a relative difference of the $y$-axis of $\{+0.6, -0.6\}$ compared to the previous step.
We use a variance of $1e-3$ for each Gaussian.
The right side of Figure \ref{app_fig:grid_walker_introduction} shows $100$ random samples for $d=5$.
The ground truth reward has $2^d$ distinct modes, each corresponding to one combination of going `up' or `down' $d$ times.
All solutions have equal probability due to the symmetry of the task.
\par 

\begin{figure}[t]
    \centering
    \includegraphics[width=0.6\textwidth]{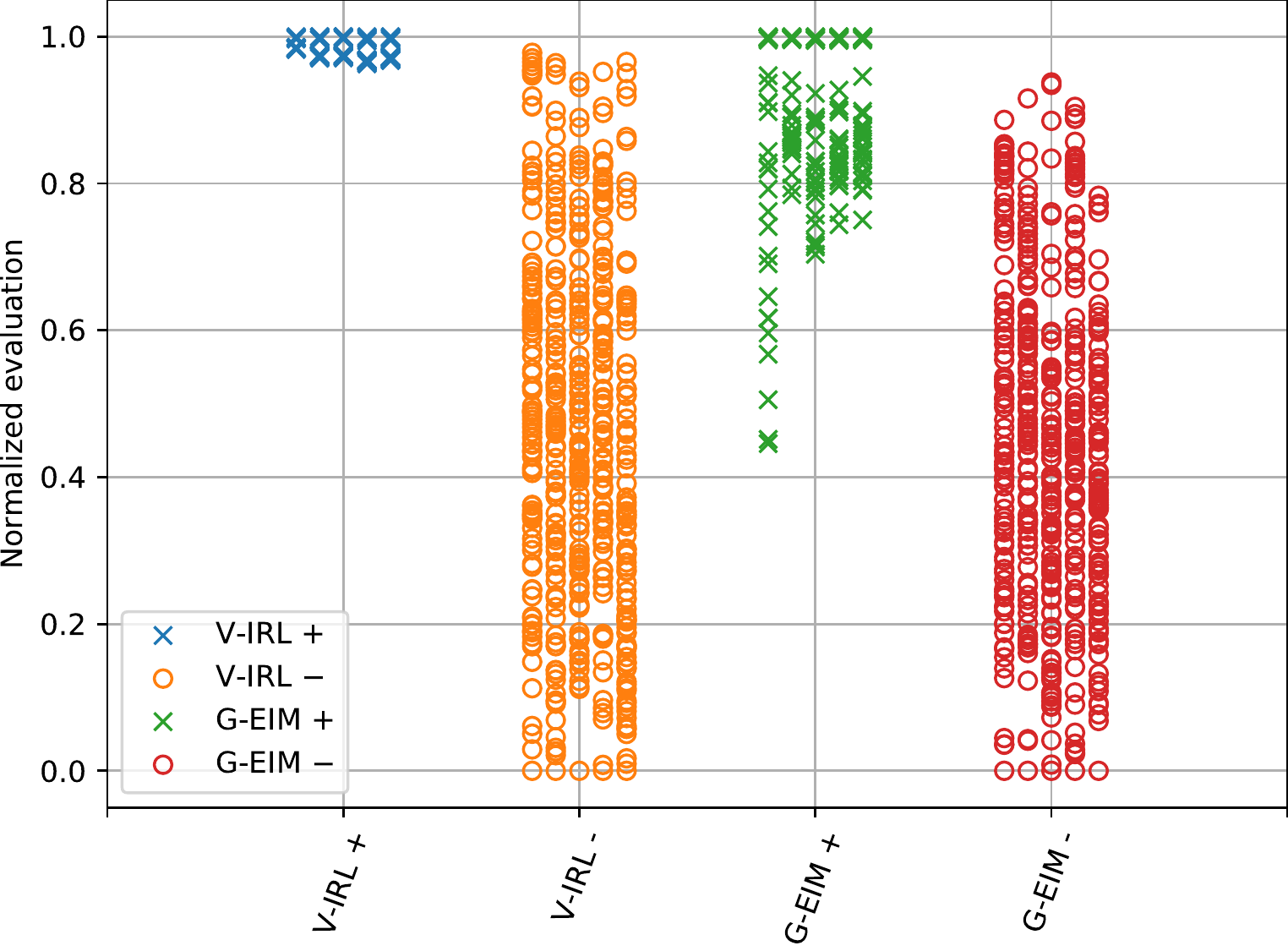}
    \caption{Normalized evaluation of the recovered reward on the grid walker task for $d=5$ for V-IRL and G-EIM. The `$+$'/`$\times$' denotes mode centers, `$-$'/`$\circ$' randomly drawn samples. The $i$-th column of the same color corresponds to the $i$-th random trials. While V-IRL represents all target modes due to its KDE, G-EIM represents those explored by its sampling distribution and only approximates the others.}
    \label{fig:grid_walker_sample_representation}
\end{figure}

\paragraph{Reward Comparison}
To explain the results of Figure \ref{fig:inference_grid_walker_vis}, we analyse how well target modes that are unexplored by the sampling policy of V-IRL and G-EIM are represented by their recovered rewards.
The steps have length $1$ and two consecutive target lines a distance of $0.8$.
The centers of the modes are therefore given by
\begin{equation*}
    \chi^+=\{-\cos^{-1}(0.8), \cos^{-1}(0.8)\}^d\text{.}
\end{equation*} 

We use this to compare the evaluations of a recovered reward at the center of each mode with evaluations at random points.
If the recovered reward represents these unexplored modes well, it will evaluate to higher values for the modes than it will for random samples.
We evaluate $5$ trials of V-IRL and G-EIM with sampling policies with $10$ components each.
We compare the $32$ target centers and $100$ negative samples randomly drawn from $\left[-1.2\cos^{-1}(0.8), 1.2\cos^{-1}(0.8)\right]^d$.
The normalized results are shown in Figure \ref{fig:grid_walker_sample_representation}.
V-IRL is able to represent both explored and unexplored mode centers well.
Opposed to this, G-EIM clearly prefers explored modes, failing to distinguish between them and unexplored ones.
We hypothesize that this is due to V-IRL making use of the Kernel Density Estimate to roughly model the full reward distribution. 
While G-EIM also uses the Kernel Density Estimation, its structure causes expert demonstrations that are unexplored by the learner policy to be less important during the training of its reward function.
This can be seen in the following equation 
\begin{equation*} 
\phi_t\left(\x\right) = \log \dfrac{\exp(R_t\left(\x\right))}{q_t\left(\x\right)},
\end{equation*}
which states that the evaluation of G-EIM for a sample $\x$ negatively depends on the log density of its sampling policy.
Linking this to the introductory example, V-IRL is able to provide paths for the path-planning task even if these paths were not explored by its sampling policy during training. If the explored paths were to become unavailable, the others would still provide valid solutions.

\section{Hyperparameters}
\label{app_sec:hyperparameters}
All discriminators are feedforward neural networks and implemented using Tensorflow \citep{tensorflow2015whitepaper} and Keras \citep{chollet2015keras}. The networks are trained using Adam \citep{kingma2015adam}, and we employ employing Dropout \citep{srivastava2014dropout}, L2 regularization and Batch Normalization \citep{ioffe2015batchnorm} to avoid overfitting.
\par 

Each task uses $8000$ expert demonstrations that are drawn from the ground truth reward of the task using Elliptical Slice Sampling \citep{nishihara2014parallel}. We train on $8000$ expert demonstrations for each task. Policy updates are performed using default hyperparameters for, and we resort to Optuna \citep{optuna_2019} for optimizing the discriminator architecture for all methods. 
We only optimize the number of policy update steps for V-IRL, as the other methods have shown to be very unstable for larger numbers of update steps.
Hyperparameter ranges are given in Table \ref{tab:app_optuna_hyperparameters}.

\begin{table}[ht!]
\caption{Ranges of the optimized hyperparameters. The parameters are optimized independently for each task and method, unless noted otherwise.}
\vspace*{0.5cm}
\centering
\label{tab:app_optuna_hyperparameters}
\begin{tabular}{llll}
\toprule
 Description & Range & Usage  \\
\midrule
Network layers & $\left[2, 4\right]$ & Neural Network \\
Layer size & $2^{\left[3, 8\right]}$ &  Neural Network  \\
Batch normalization & $\left[\text{False}, \text{True}\right]$ & Neural Network \\
Learning rate & $\left[5.0e-5, 1.0e-3\right]$ & Neural Network  \\
Dropout & $\left[0, 0.5\right]$ & Neural Network \\
L2-Norm & $\left[0, 1\right]$ & Neural Network \\
Bandwidth & $\left(0, 1\right]$ & KDE \\
Policy update steps & $\left[1, 200\right]$ & Learner Policy\\
\bottomrule
\end{tabular}
\end{table}
\par 

\paragraph{Chosen Hyperparameters}
\label{app_sec:chosen_hyperparameters}
This section lists the hyperparameters chosen by Optuna for all experiments and optimized methods.

\begin{table}[ht!]
\caption{Hyperparameters used for the random Gaussian experiments. Values with a
`$*$' are not optimized and instead set to a default value. We fix the bandwidth according to Silverman's rule 
\citep{silverman1986density}
for simplicity.}
\vspace*{0.5cm}
\centering
\label{tab:app_optuna_hyperparameters_2d}
\begin{tabular}{lllllll}
\toprule
Parameters & V-IRL 1c & V-IRL 10c & G-EIM 1c & G-EIM 10c & EIM 1c & EIM 10c\\
\midrule
Network Layers & $4$ & $4$ & $4$ & $3$ & $4$ & $4$\\
Layer size & $256$ & $256$ & $256$ & $256$ & $256$ & $256$\\
Batch norm & $True$ & $False$ & $True$ & $False$ & $False$ & $True$\\
Learning Rate & $2.68e-4$ & $3.81e-4$ & $1.09e-4$ & $6.22e-4$ & $8.43e-4$ & $3.28e-4$\\
Dropout & $0.01$ & $0.01$ & $0$ & $0$ & $0.02$ & $0.02$\\
L2-Norm & $0$ & $5.96e-8$ & $0$ & $4.77e-7$ & $2.98e-8$ & $9.54e-7$\\
Bandwidth & $0.215^*$ & $0.215^*$ & $0.215^*$ & $0.215^*$ & $\text{\sffamily X}$& $\text{\sffamily X}$\\
Update steps  & $1$ & $162$ & $5^*$ & $5^*$ & $1^*$ & $1^*$\\
\bottomrule
\end{tabular}
\end{table}

\begin{table}[ht!]
\caption{Hyperparameters used for the grid walker experiments. Values with a
`$^*$' are not optimized and instead set to a default value.}
\vspace*{0.5cm}
\centering
\label{tab:app_optuna_hyperparameters_walker}
\begin{tabular}{lllllll}
\toprule
Parameters & V-IRL & G-EIM\\
\midrule
Network Layers & $3$ & $3$\\
Layer size & $256$ & $64$\\
Batch norm & $True$ & $False$\\
Learning Rate & $4.25e-4$ & $4.38e-4$\\
Dropout & $0.07$ & $0$\\
L2-Norm & $4.77e-7$ & $2.44e-4$\\
Bandwidth & $0.136$ & $0.864$\\
Update Steps & $63$ & $5^*$\\
\bottomrule
\end{tabular}
\end{table}

\end{document}